\documentclass{article}

\usepackage{arxiv}
\usepackage{lineno}

\usepackage{times}
\usepackage{epsfig}
\usepackage{graphicx}
\usepackage{amsmath}
\usepackage{amssymb}
\usepackage{booktabs}
\usepackage{yansong}

\def \a {\alpha}

\def \E { \mathbb{E} }

\def \d { \mathrm{d} }

\usepackage{algorithm}
\usepackage{algpseudocode}

\usepackage[breaklinks=true,bookmarks=false]{hyperref}

\begin{document}
\nolinenumbers

\title{Fast Diffusion Probabilistic Model Sampling through
the lens of Backward Error Analysis}


\author{Yansong Gao$^{1, *}$, Zhihong Pan$^2$, Xin Zhou$^2$, Le Kang$^2$, Pratik Chaudhari$^1$\\
\and
$^1$University of Pennsylvania, $^*$Work done during an internship, $^2$Baidu Research USA
}

\maketitle

\begin{abstract}
  Denoising diffusion probabilistic models (DDPMs) are a class of powerful generative models. The past few years have witnessed the great success of DDPMs in generating high-fidelity samples. A significant limitation of the DDPMs is the slow sampling procedure. DDPMs generally need hundreds or thousands of sequential function evaluations (steps) of neural networks to generate a sample. This paper aims to develop a fast sampling method for DDPMs requiring much fewer steps while retaining high sample quality. The inference process of  DDPMs approximates solving the corresponding diffusion ordinary differential equations (diffusion ODEs) in the continuous limit. This work analyzes how the backward error affects the diffusion ODEs and the sample quality in DDPMs. We propose fast sampling through the \textbf{Restricting Backward Error schedule (RBE schedule)} based on dynamically moderating the long-time backward error. Our method accelerates DDPMs without any further training. Our experiments show that sampling with an RBE schedule generates high-quality samples within only 8 to 20 function evaluations on various benchmark datasets. We achieved 12.01 FID in 8 function evaluations on the ImageNet $128\times128$, and a $20\times$ speedup compared with previous baseline samplers.
\end{abstract}

\section{Introduction}
Denoising Diffusion Probabilistic Models (DDPMs)~\cite{sohl2015deep, ho2020denoising} have emerged as a powerful class of generative models. DDPMs model the data distribution through an iterative denoising process. Recently, DDPMs have achieved impressive performances on a variety of applications, including unconditional and conditional image generation~\cite{song2019generative, nichol2021improved, dhariwal2021diffusion, ho2022classifier}, video generation~\cite{ho2022video}, 3D point cloud generation~\cite{lyu2021conditional}, text to speech~\cite{chen2020wavegrad, kong2020diffwave}, and image super-resolution~\cite{saharia2021image, li2022srdiff}. Despite the remarkable success, inference of DDPMs ( e.g., sampling and density estimation) takes much longer to produce high-quality results compared with other generative models such as GANs~\cite{goodfellow2014generative}. To give a few examples, DDPMs may take $4000$ steps~\cite{nichol2021improved} to generate one sample. Each step requires evaluating the learned neural network once. Such inefficiency becomes a critical bottleneck for the applications of DDPMs. For this reason, there is an urgent request to design fast samples for DDPMs. 

Several works aim at speeding up the inference process for DDPMs. One category of methods includes knowledge distillation~\cite{salimans2022progressive, luhman2021knowledge}, learning sampling trajectory, and searching for better noise schedule~\cite{san2021noise, lam2021bilateral, nichol2021improved, watson2021learning}. Such methods require a training stage before the sampling stage. Hence their applicability and effectivity might be limited when adapting their methods to different models, datasets, and the number of sampling steps. The second category of methods modifies/optimize the forward/backward noising process~\cite{song2020score} in a training-free manner, including dynamic programming~\cite{watson2021learningdp} and analytical generation process~\cite{bao2022analytic}. An important and effective instance is the Denoising Diffusion Implicit Model (DDIM)~\cite{song2020denoising}, a non-Markovian noising process. Another category of training-free methods modifies the numerical solver for stochastic differential equations (SDEs) or ordinary differential equations (ODEs) associated with the DDPMs~\cite{jolicoeur2021gotta, tachibana2021taylor}. In particular, ~\cite{song2020score} solves a marginal equivalent ODE known as the Probability Flow (PF) using a black box ODE solver. ~\cite{liu2022pseudo, lu2022dpm, zhang2022fast} compute DDIM with high order methods and achieve further acceleration. The deterministic DDIM can also be viewed as a time discretization of the PF as it matches the latter in the continuous limit~\cite{song2020denoising}.

This paper develops a simple and robust fast sampling method for DDPMs with much fewer steps while retaining high sample quality. Our approach is inspired by backward error analysis, a method for the numerical analysis of ordinary differential equation (ODE) solvers~\cite{hairer2003geometric}. Backward error analysis is a powerful tool for studying the long-time behavior of numerical ODE solvers. The critical insight of backward error analysis is that we can describe the bias introduced when integrating an ODE with finite step sizes by introducing a modified ancillary flow. This modified flow is derived to ensure that discrete iterates of the original ODE lie on the path of the continuous solution to the modified flow. 

Our contributions are summarized as follows: 1) The inference process of  DDPMs approximates solving the corresponding diffusion ordinary differential equations (diffusion ODEs) in the continuous limit~\cite{song2020score}. This work analyzes how the backward error affects the diffusion ODEs and the sample quality in DDPMs; 2) we propose fast sampling through the Restricting Backward Error schedule (RBE schedule) based on dynamically moderating the long-time backward error, our method accelerates DDPMs without any further training; 3) the experiments show that sampling with an RBE schedule generates high-quality samples within only 8 to 20 function evaluations on various benchmark datasets. We achieved 12.01 FID in 8 function evaluations on the ImageNet $128\times128$, and a $20\times$ speedup compared with previous baseline samplers.
 


\begin{figure}[t]
\begin{center}
   \includegraphics[width=0.6\linewidth]{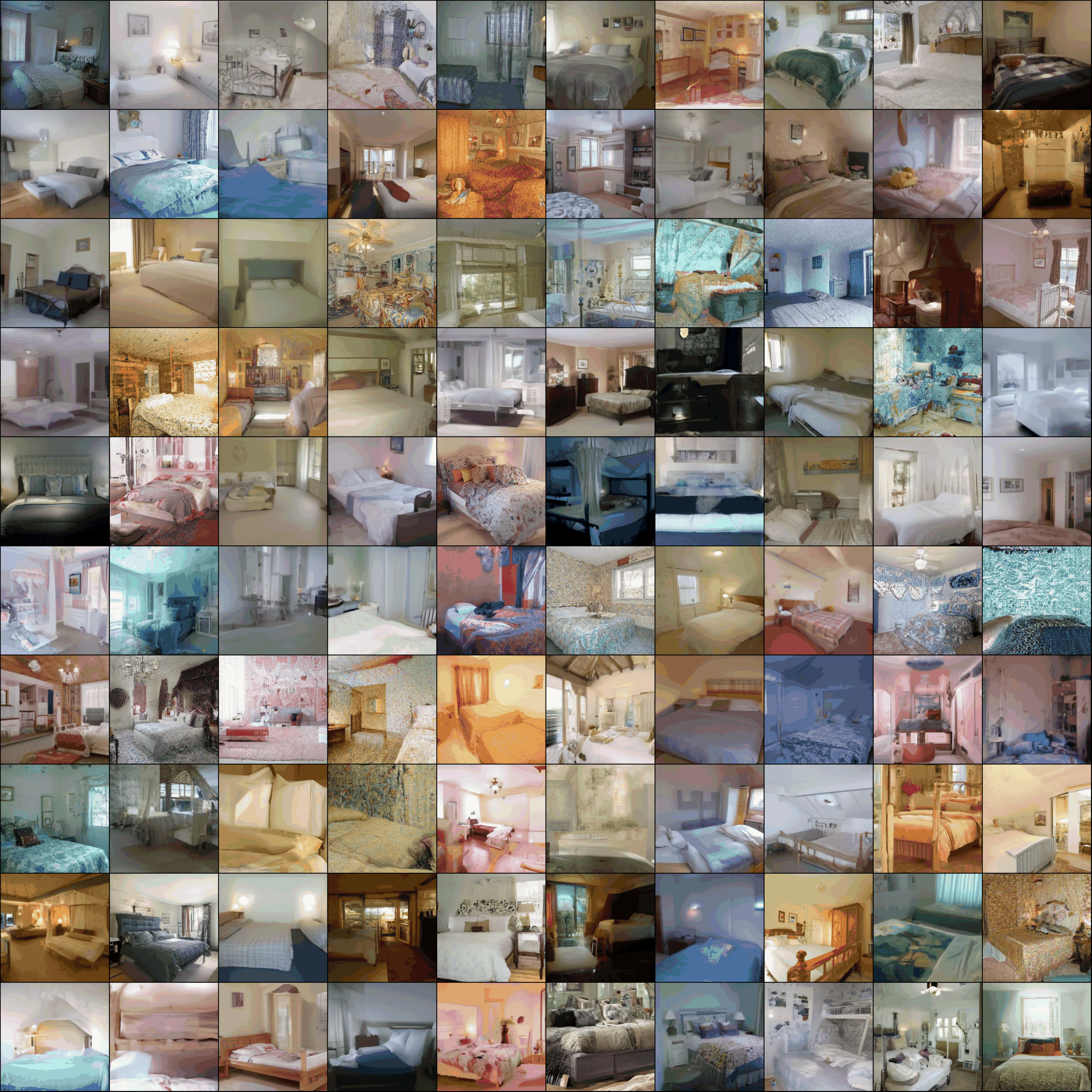}
\end{center}
   \caption{ Sampling through RBE schedule within $10$ forward evaluations. We use the diffusion model~\cite{nichol2021improved} pre-trained on on LSUN bedroom 256×256.}
\label{fig: lsun_bed}
\end{figure}

\section{Denoising Diffusion Probabilistic Models}

Denoising Diffusion Probabilistic Models (DDPMs)\cite{sohl2015deep, ho2020denoising} are defined in terms of a forward Markovian diffusion process $q$ and a learned reverse process $p$. The forward diffusion process gradually adds Gaussian noise to a data point $x_0$ through $N$ iterations, 
\beqs{
q(x_{1:N} \mid x_0) = \prod_{i = 1}^{N} q( x_{i} \mid x_{i-1} ), 
}
and 
\aeq{
q( x_i \mid x_{i-1} ) &= \mathcal{N}( x_i \mid \sqrt{ \a_i} x_{i-1}, ( 1 - \a_i ) \mathbf{I }) \nonumber \\
& = \mathcal{N}( x_i \mid \sqrt{ \f{\g_i}{\g_{i-1}}} x_{i-1}, \rbr{ 1 - \f{\g_i}{\g_{i-1}} } \mathbf{I}), 
\label{eq: ddpm_forward}
}
where $\g_0 = 1,\ \g_i = \prod_{j=1}^i \a_j$. The scalar parameters $\a_{1:N}$ determine the variance of the noise added at each diffusion step, subject to $0 < \a_i < 1$. The learned reverse process aims to model $q(x_0)$ by inverting the forward process, gradually removing noise from the signal starting from isotropic Gaussian noise $x_N$,
\aeq{
&x_N \sim \mathcal{N}( 0, I),\nonumber \\
&p_\theta ( x_{0:N} ) = p( x_N ) \prod_{i = 1}^N p_\theta ( x_{i-1} \mid x_i ), \nonumber \\
&p_\theta ( x_{i-1} \mid x_i ) = \mathcal{N}( x_{i-1} \mid \mu_\theta(i, x_i), \s_i^2 I ).
}
The parameters $\theta$ of the reverse process can be optimized by maximizing the following variational lower bound on the training set~\cite{ho2020denoising}, 
\aeq{
L_{i-1} &= \E_{x_0, \e} \left[ \left\| \sqrt{\f{ 1}{\a_i}} \rbr{ x_i(x_0, \e) - \f{1 - \a_i}{\sqrt{1 - \g_i}} }\e \right. \right.  \nonumber \\
& \left. \left.\quad - \mu_\theta(i, x_i(x_0,\e) )\right\|^2 \right] \nonumber \\
& = \E_{x_0, \e} \sbr{ \f{ (1 - \a_i)^2 }{\a_i(1 - \g_i)}\left\|  \e - \e_\theta( i, x_i(x_0, \e) ) \right\|^2 },
}
where $\g_i = \prod_{j=1}^i \a_j$, $x_i(x_0, \e) = \sqrt{ \g_i } x_0 + \sqrt{1 - \g_i}\e$, and $\e \sim \mathcal{N}(0, I)$. $\e_\theta$ is an estimate of the noise $\e$. The relationship between $\e_\theta$ and $\mu_\theta$ is $\mu_\theta = \f{1}{\sqrt{\a_i}}\rbr{x_i - \f{1 - \a_i}{\sqrt{1- \g_i}}\e_\theta}$. 

\subsection{Diffusion SDEs}
Before introducing our approach to efficiently sampling from DDPMs, it is helpful to link DDPMs to continuous time affine diffusion processes, as it reveals the intuition behind our approach. First we switch the domain of index $i$ from $\{0, 1,2,...,N \}$ to continuous interval $[0, T]$, in which $t_i = \f{i T}{N}$ and each $x_i$ is corresponding to the value at moment $t_i$, e.g. $x_i = x_{t_i}$. As $N$ goes to infinity, $\{ x_{t_i} \}_{i=0}^N$ and $\{\g_{t_i} \}_{i=0}^N$ approach continuous functions $x_t$ and $\g_t$ on interval $[0, T]$. Let $x_0 \sim q(x_0)$ denote a data point drawn from the empirical distribution of the train set. Let $q(x_t \mid x_0)$ denote a stochastic process for $t \in [0, T]$ defined through the following stochastic differential equation (SDE),
\aeq{
\df x_t 
 = \f{x}{2}\f{\df \log \g_t}{ \df t}  \df t + \sqrt{ - \f{\df \log \g_t}{ \df t} } \df w, 
\label{eq: ddpm_sde}
}
where $\e \sim \mathcal{N}(0, I)$, $\df w$ is Wiener process. Equation~\ref{eq: ddpm_forward} matches the differential Equation~\ref{eq: ddpm_sde} in the continuous limit. Meanwhile, the reverse time denoising process also coincides with a time-continuous stochastic differential Equation~\cite{song2020score} (diffusion SDE) starting from time $T$ to $0$,
\beq{
\df x_t = \f{\df \log \g_t}{\df t} \rbr{ \f{x}{2} - \f{\e_\theta( t, x_t )}{\sqrt{ 1 - \g_t}} }\df t + \sqrt{ \f{\df \log \g_t}{ \df t} }\df w.
\label{eq: diffSDE}
}

Equation~\ref{eq: diffSDE} is Variance Preserving stochastic differential equations (VP-SDEs)~\cite{song2020score}. We can generate samples by solving the diffusion SDE with numerical solvers. This approach discretizes the diffusion SDE from $T$ to $0$. \cite{song2020score} proved that the classical ancestral sampling method for DDPMs is a first-order SDE solver for Equation~\ref{eq: diffSDE}. However, these first-order methods usually require hundreds of or thousands of function evaluations to converge~\cite{song2020score, ho2020denoising}, resulting in extremely slow sampling speed.

\subsection{Diffusion ODEs (Probability Flow) }
\label{sec: diffODE}

By randomness of the Wiener process, the first-order SDE solvers always require a limited step size~\cite{kloeden1992stochastic}. Simulating SDEs with a large step size (small number of steps) often results in non-convergence, especially in high-dimensional spaces. For fast sampling, ~\cite{song2020score} shows that Equation~\ref{eq: diffSDE} has a marginal equivalent ODE known as the diffusion ODE (Probability Flow),
\beq{
\df x_t = \f{\df \log \g_t}{\df t} \rbr{ \f{x}{2} - \f{\e_\theta( t, x_t )}{2\sqrt{ 1 - \g_t}} }\df t .
\label{eq: diffODE}
}
We can draw samples by solving the diffusion ODE from $T$ to $0$. The deterministic DDIM can also be viewed as a time discretization of the diffusion ODEs as it matches the latter in the continuous limit~\cite{song2020denoising}. In contrast with SDEs, ODEs can be solved with larger step sizes as there is no Wiener process. Diffusion ODEs show us a new possibility to accelerate the reverse process. Furthermore, we can use efficient numerical ODE solvers to accelerate the sampling. 

\cite{song2020score} solves the diffusion ODEs using the RK45 ODE solver~\cite{dormand1980family}. This approach generates samples in 60 function evaluations on the CIFAR-10 dataset~\cite{krizhevsky2009learning} and achieves comparable quality with a 1000-step SDE solver for diffusion SDEs. \cite{lu2022dpm, zhang2022fast} apply change of variable and solve diffusion ODEs using the numerical methods called exponential integrators~\cite{ostermann2010exponential}. Their approaches generate satisfactory samples in the few-step (10 steps) sampling regime. However, \cite{lu2022dpm} fails dramatically in the fewer-step (8 steps) sampling regime (Section~\ref{sec: exp}). We know there is still a lack of training-free samplers for DDPMs in the few-step sampling regime. In addition, compared to the commonly-used GANs, diffusion ODE solvers are still not fast enough for real-time applications.

\section{A Backward Error Analysis of Diffusion ODEs}

As we have discussed in Section~\ref{sec: diffODE}, simulating SDEs with discrete updates is generally difficult in high dimensions~\cite{kloeden1992stochastic}, and it is hard to converge within a few steps. In contrast, solving ODEs is relatively easier. This fact yields a potential for fast samplers and motivates us to design an efficient solver for diffusion ODEs to enable fast and high-quality few-step sampling. 

The critical insight of this work is inspired by Backward error analysis. Backward error analysis has great potential to clarify the role of finite step size and to help identify the implicit biases of different ODE solvers. We, therefore, give a detailed introduction to the core methodology in Section~\ref{sec: intro_BEA} before deriving our diffusion ODE solvers in Section~\ref{sec: alg} and Section~\ref{sec: RBE}. 


\subsection{ An Introduction to Backward Error Analysis}
\label{sec: intro_BEA}
In numerical analysis, we often wish to integrate ODEs of the form $\f{\df x}{\df t} = f(t, x)$. This system usually
cannot be solved analytically, forcing us to approximate the continuous flow with discrete updates, e.g., the Euler step 
\beq{
x_{t+h} = x_t + h \cdot f(t, x_t).
\label{eq: euler}
}
However, discrete updates introduce approximation error when the step size $h$ is finite. In order to study the bias introduced by this approximation error, we assume the step size $h$ is relatively small, and introduce a modified flow $\f{\df x}{\df t} = \Tilde{f}(t, x)$, 
\beq{
\Tilde{f}(t, x) = f(t, x) + h f_1( t, x ) + h^2 f_2( t, x ) +...,
\label{eq: modified_flow1}
}
where $f_1, f_2...$ are the correction terms. The modified flow of $\Tilde{f}(t, x)$ converges to the original flow of $f(t, x)$ as $h$ goes to $0$, but it differs from the original ODE if $h$ is finite. The goal of backward error analysis is to choose the correction terms $f_1(t, x)$ such that the iterates obtained from discrete updates of the original flow with small finite step sizes lie on the path given by the continuous solution to the modified flow with vanishing step sizes. 

The standard derivation of backward error analysis follows by taking a Taylor expansion in $h$ of the
solution to the modified flow $x_{t+h}$ (see~\cite{hairer2003geometric}). We obtain the derivatives of $x_{t +h }$ recursively using the modified flow equation $\f{\df x}{\df t} = \Tilde{f}(t, x)$, and we identify the correction terms $f_i(t, x)$ by ensuring this Taylor expansion matches the discrete update (e.g., $x_{t +h} = x_t + h f(t, x_t) $ for all powers of $h$. This section will build an intuition for the origin of the corrections terms. We will clarify how we might apply this analysis to high-quality few-step sampling. First, we will identify the path of the continuous modified flow by solving it analytically in the continuous limit. Then we will compare this continuous path to the single Euler step $x_{t+h} = x_t + h \cdot f(t, x_t)$. Imagine tracking the modified flow $\Tilde{f}(t, x)$ with time step size $h$,
\aeq{
x_{t+h } &= x_t + \int_{t}^{t+h } \Tilde{f}(s, x_s) \df s  \nonumber \\
& = x_t + h \Tilde{f}(t, x_t) + \f{h^2}{2} \f{\df \Tilde{f}(t, x_t)}{ \df t} + O(h^3). 
\label{eq: modified_flow2}
}
We arrived at Equation~\ref{eq: modified_flow2} by taking the Taylor expansion of $\Tilde{f}$ then integrating over time. Here we assume $\f{\df \Tilde{f}}{ \df t}$ does exist. Next, we replace $\Tilde{f}$ with its definition in Equation~\ref{eq: modified_flow1}. As we will see below,
\aeq{
x_{t+h } 
& = x_t + h f(t, x_t) + h^2 \rbr{ f_1( t, x_t ) + \f{1}{2}\f{\df f(t, x_t)}{ \df t} } + O(h^3). 
\label{eq: modified_flow3}
}
Equation~\ref{eq: modified_flow3} is the key component of backward error analysis. It describes the path taken while integrating the
continuous modified flow $\Tilde{f}$ over a discrete time step of length $h$. 

Notice that we have assumed that the Taylor expansion in Equation~\ref{eq: modified_flow3} converges, while the higher
order terms $O(h^3)$ will contain higher order derivatives of the original flow $f$. Backward error
analysis implicitly assumes that $f$ is an analytic function in $t$ and $x$. We refer the reader to~\cite{hairer2003geometric} for a detailed introduction. 

\heading{First order correction:} we will now derive the first order correction $f_1$ of the
modified flow $\Tilde{f}$ in Equation~\ref{eq: modified_flow1}. Recall that the discrete updates follow by $x_{t+h} = x_t + h f(t, x_t)$. In order to make sure that the continuous modified flow coincides with the discrete updates, we need all terms at $O(h^2)$ and above in Equation~\ref{eq: modified_flow3} to vanish. As for the second order terms, it implies that $f_1( t, x_t ) + \f{1}{2}\f{\df f(t, x_t)}{ \df t} = O(h)$, and yields the first order correction,
\aeq{
f_1(t, x) 
 =  - \f{1}{2} \rbr{ \f{\partial f(t, x)}{\partial t}  + \f{1}{2}\nabla  \| f(t, x)  \|^2   },
}
where $\nabla$ is the gradient with respect to $x$.

\begin{figure}[t]
\begin{center}
   \includegraphics[width=0.5\linewidth]{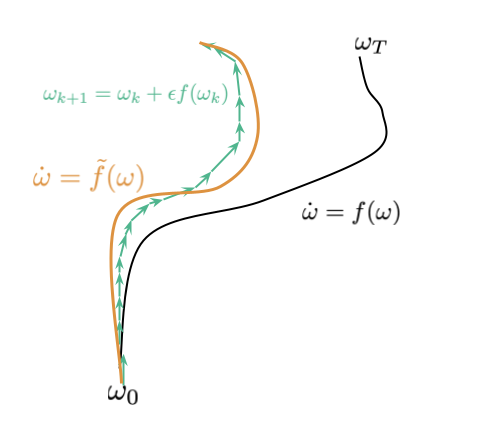}
\end{center}
   \caption{The cartoon illustration for backward error analysis. Iterates obtained from discrete
Euler updates $(w_{k+1} = w_k + \e f(w_k))$ of the original flow $(\dot{w} = f(w))$ with small finite step sizes $\e$ lie
close to the path given by the continuous solution to the modified
flow $(\dot{w} = \tilde{f}(w))$ with vanishing step sizes.}
\label{fig: modified_flow}
\end{figure}

We conclude that if the time step $h$ is sufficiently small such that we can neglect higher order terms in Equation~\ref{eq: modified_flow3}, then the discrete Euler updating path Equation~\ref{eq: euler} lies close to the path of the modified ODE (see Fig.~\ref{fig: modified_flow}),
\beq{
\f{\df x}{\df t} =f(t, x) - \f{h_t}{ 2} \rbr{ \f{\partial f(t, x)}{\partial t}  + \f{1}{2}\nabla  \| f(t, x)  \|^2   },
\label{eq: modified_flow4}
}
where $h_t$ is the finite step size at moment $t$.

\subsection{ Dynamically Restricting the Backward Error}
\label{sec: alg}
We now derive our first algorithm. Recall that the flow of the diffusion ODEs satisfies,
\beq{
\f{\df x}{ \df t} = \f{\df \g_t}{ \df t} \rbr{ \f{x}{2 \g_t} - \f{\e_{\theta}( t, x )}{ 2 \g_t \sqrt{1 - \g_t}}}. 
}
By applying change-of-variable, we switch to consider the following ODE,
\beq{
\f{\df x}{ \df \g} = f_\theta( \g, x):= \f{x}{2 \g} - \f{\e_{\theta}( \g, x )}{ 2 \g \sqrt{1 - \g}},
\label{eq: ddim_ode}
}
where $\g$ is between $0$ and $1$, and $\e_{\theta}( \g_t, x ) = \e_{\theta}( t, x )$. The deterministic DDIM can be viewed as a discrete version of Equation~\ref{eq: ddim_ode} as it matches the latter in the continuous limit~\cite{song2020denoising}. Followed by Equation~\ref{eq: modified_flow4}, the discrete flow lies close to the path of the modified ODE,
\aeq{
\f{\df x}{\df \g} &= \Tilde{f_{\theta}}( \g, x ) \nonumber \\
&= f_{\theta}( \g, x) - \f{h_\g}{ 2} \rbr{ \f{\partial f_{\theta}(\g, x)}{\partial \g}  + \f{1}{2}\nabla  \| f_{\theta}(\g, x)  \|^2   }, 
}
where $h_\g$ is the finite step size at the moment $\g$. If the modified flow $\Tilde{f_{\theta}}$ coincides with the original Diffusion ODE $f_{\theta}$ at $\g$, as the moment goes from $\g$ to $\g + h_{\g}$, the discrepancy between the modified flow and the original flow approximately becomes
\beq{
\f{h^2_\g}{ 2} \rbr{ \f{\partial f_{\theta}(\g, x)}{\partial \g}  + \f{1}{2}\nabla  \| f_{\theta}(\g, x)  \|^2   }.
\label{eq: discrepency}
}
Observing the discrepency~\ref{eq: discrepency} motivates us to design a solver for diffusion ODEs to enable high-quality few-step sampling. Our first algorithm dynamically crafts the step size $h_\g$ to maintain restricting the backward error. Below we present our first method and name it as sampling through \emph{Dynamically Restricting the Backward Error (DRBE)} schedule.

\begin{algorithm}
\caption{Improving the sampling quality by dynamically restricting the backward error}
\begin{algorithmic}
\Require threshold $r > 0$; $f_{\theta}(\cdot, \cdot)$; initial state $x_{T}$.
\State $t \gets T$
\State $x \sim \mathcal{N}(0, I)$
\State $\g \gets  \g_t$
\State $sche \gets \{ t\}$
\While{$t > 0 $}
    \State $h \gets - \min\cbr{ 1 - \g,\  \sqrt{\f{r}{ \left| \f{ \partial f_\theta( \g, x )}{ \partial \g} + \f{1}{2}\nabla \| f_\theta( \g, x )\|^2 \right|}}} $
    \State $x \gets x + h f_{\theta}(\g, x) $ 
    \State $\g \gets \g + h $
    \State $t \gets t^{-1}(\g)$
    \State $sche \gets sche \cup \{ t\}$
\EndWhile\\
\Return $x, sche$
\end{algorithmic}
\label{alg: adp}
\end{algorithm}

In Algorithm~\ref{alg: adp}, $t^{-1}(\g)$ is the inverse function of $\g_t$. $t^{-1}(\cdot)$ has an analytical formulation for both linear and cosine noise schedules used in~\cite{nichol2021improved, ho2020denoising}. The crafted threshold $r>0$ governs the sample quality and sampling speed trade-offs. The average inference steps increase/decrease if we accept a small/large threshold. We carefully fine-tune the threshold to achieve high-quality sampling within 8, 10, 12, 15, and 20 steps.

\begin{remark}
Followed by Equation~\ref{eq: discrepency}, the step size $h_\g$ is governed by the magnitude of the correction term $\f{\partial f(\g, x)}{\partial \g}  + \f{1}{2}\nabla  \| f(\g, x)  \|^2 $. The correction term encodes the second-order derivatives information of the diffusion ODEs. Intuitively, in numerical analysis, we can simulate the ODEs with small/large step sizes while the curvature of the flow is significant/minor. 
\end{remark}

\subsection{Restricting the backward error learns effective inference schedule}
\label{sec: RBE}
As shown in the previous section, sampling through DRBE schedule requires one function evaluation and one gradient evaluation per step. Despite the more expensive steps, we observe a fun fact in practical experiments: even though the initial states $x$ are randomly sampled from isotropic Gaussian distribution, the output $sche$ in Algorithm~\ref{alg: adp} is roughly unchanged if we fix the inference model $\theta$. We suspect that this phenomenon may stem from the flow of diffusion ODEs being locally parallel. This observation motivates us to design a cheaper but equally effective approach for fast sampling compared with DRBE schedule. Suppose we have generated a small batch of $N$ samples using Algorithm~\ref{alg: adp} within $K$ inference steps. Our second method first learns a newly crafted inference schedule by restricting the backward error and taking the average of the collected $K-$steps schedules $\{sche^i\}_{i=1}^N$. More precisely,
for $k =1, 2,...,K+1$, we compute the following statistics
\beq{
t_k = \f{1}{N}\sum_{i=1}^N sche^i[k],
}
here $sche^i[k]$ is the $k$-th item in $sche^i$. We name $\{t_1, t_2,...,t_{K+1} \}$ as \emph{Restricting Backward Error schedule (RBE schedule)}.  Our second algorithm utilizes the learnt inference schedule $\{t_1, t_2,...,t_{K+1} \}$ and only requires one function evaluation per step, as shown in below,

\begin{algorithm}
\caption{Improving the sampling speed through the RBE schedule} 
\begin{algorithmic}
\Require $\g_t$; $f_{\theta}(\cdot, \cdot)$; RBE schedule $\{ t_1, t_2,..., t_K\}$.
\State $x \sim \mathcal{N}(0, I)$
\For{$k \gets 1$ to $K$}:
    \State $h \gets \g_{t_{k+1}} - \g_{t_{k}}$
    \State $x \gets x + h f_{\theta}(\g_{t_{k}}, x) $ 
\EndFor\\
\Return $x$
\end{algorithmic}
\label{alg: customized}
\end{algorithm}

In Section~\ref{sec: exp}, we show that sampling through the RBE schedule~\ref{alg: customized} achieves high-quality sampling with in few function evaluations.

\section{Experiments}
\label{sec: exp}

In this section, we show that generating samples through DRBE schedule~\ref{alg: adp} and RBE schedule~\ref{alg: customized} can significantly speed up the sampling of existing pre-trained DDPMs, with both linear noise models~\cite{ho2020denoising} and cosine noise models~\cite{nichol2021improved}. We vary the number of function evaluations (NFE), which is the number of calls to the noise prediction model $\e_\theta(t, x)$, and compare the sample quality among RBE schedule, DRBE schedule, and other methods. For each experiment, We draw $50K$ samples and use the widely adopted Fenchel Inception Distance (FID~\cite{heusel2017gans})
to evaluate the sample quality, lower FID usually indicates better sample quality. We refer to the Appendix for further implementation details of the experiments.

\subsection{Comparison with other training-free samplers}

We generate samples through DRBE and RBE schedules using pre-trained DDPMs and then compare our results with other DDPMs samplers, including DDPM~\cite{ho2020denoising}, DDIM\cite{song2020denoising}, and Dpm-solvers~\cite{lu2022dpm}.  
We compare the sample quality by varying NFE from $8$ to $1000$.

Specifically, we use 
the DDPMs trained by $L_{hybrid}$ in~\cite{nichol2021improved} on ImageNet 64x64~\cite{deng2009imagenet} with cosine
noise schedule; the classifier guidance DDPMs in~\cite{dhariwal2021diffusion} on ImageNet 128x128~\cite{deng2009imagenet} with linear noise schedule, whereas the classifier guidance scale is $1.0$; the discrete-time model in~\cite{nichol2021improved} on LSUN 256x256~\cite{yu2015lsun} pre-trained with linear noise schedule. For the models trained on ImageNet, we only use their “mean” model and omit the “variance” model. 

\begin{figure}[t]
\begin{center}
   \includegraphics[width=0.55\linewidth]{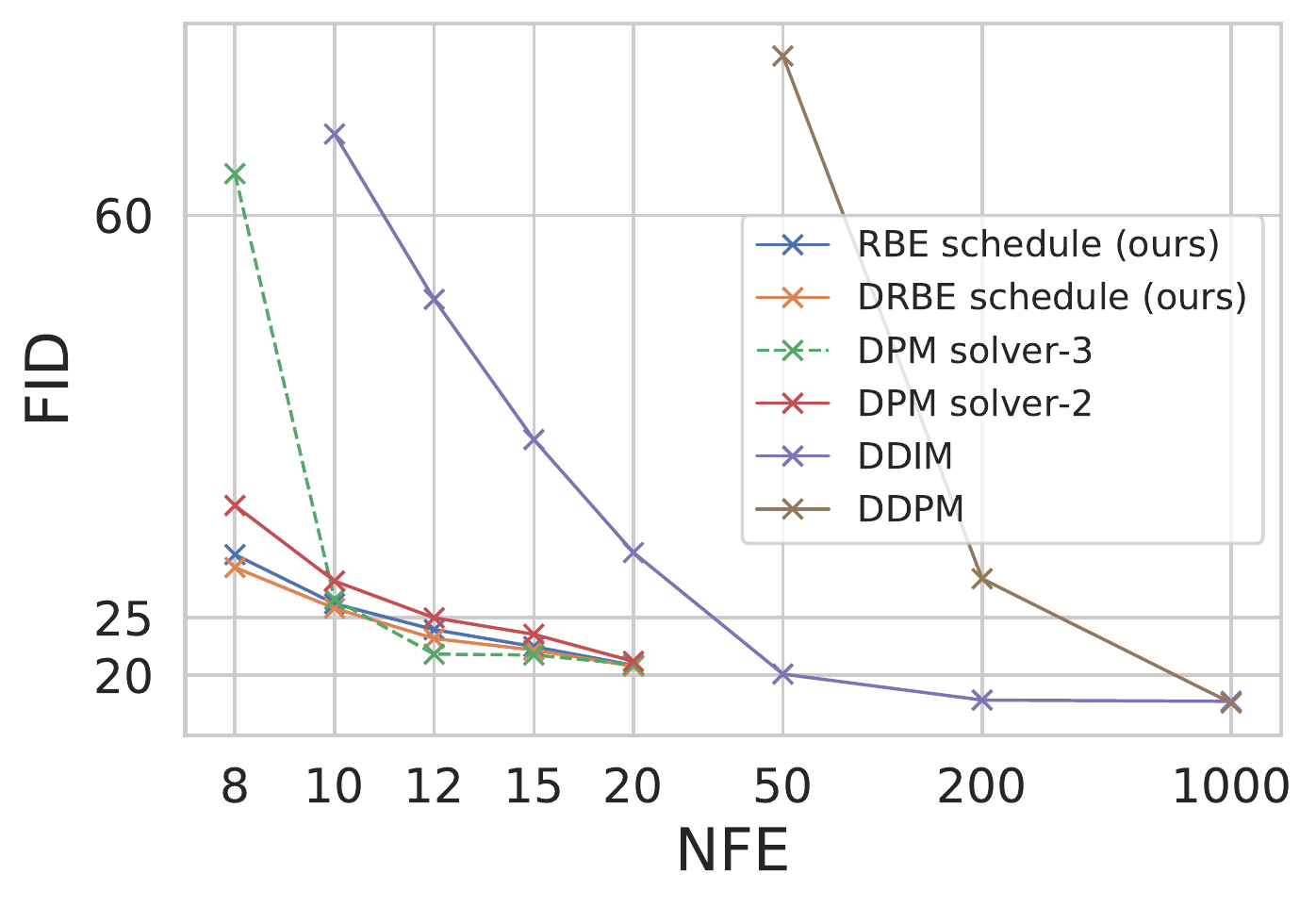}
\end{center}
   \caption{Sample quality measured by FID $\downarrow$ on ImageNet 64×64 with the diffusion model~\cite{nichol2021improved}, varying the number of function evaluations (NFE): 8, 10, 12, 15, 20, 50, 200, 1000. For DDIM and DDPM, we use uniform re-spacing time steps for all the experiments. Sampling through our RBE schedule shows uniform stability and superiority in a few inference step regimes. Our method achieves FID 25.82 in 10 function evaluations and at least 20$\times$ speedup compared with DDPM.  }
   
\label{fig: img64}
\end{figure}

\begin{figure}[t]
\begin{center}
   \includegraphics[width=0.55\linewidth]{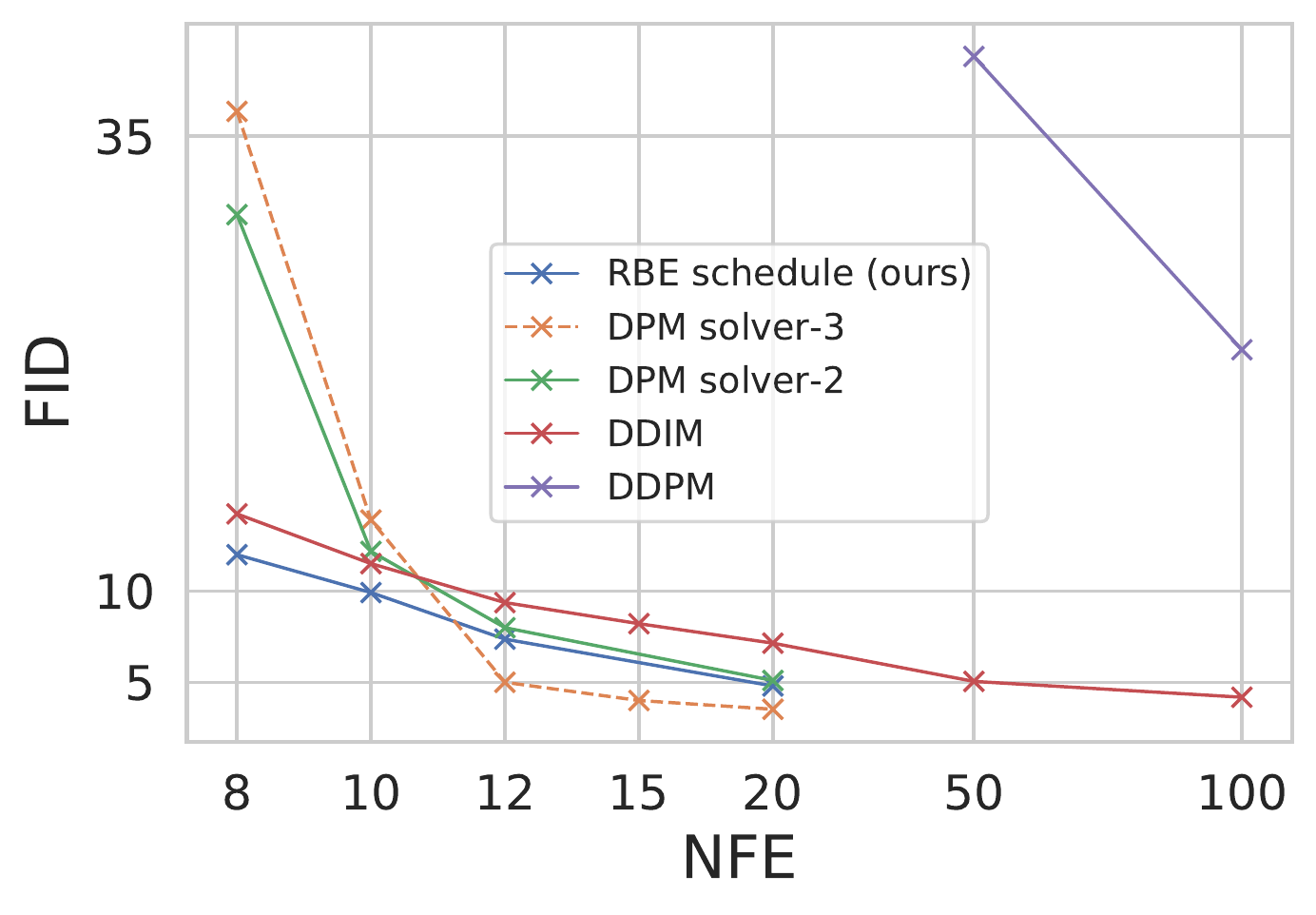}
\end{center}
   \caption{Sample quality measured by FID $\downarrow$ on ImageNet 128×128 with the diffusion model~\cite{dhariwal2021diffusion}, varying the number of function evaluations (NFE): 8, 10, 12, 15, 20, 50, 100. For DDIM and DDPM, we use uniform re-spacing time steps for all the experiments. Sampling through RBE schedule achieves FID 12.01 within only 8 function evaluations. }
\label{fig: img128}
\end{figure}

\begin{figure}[t]
\begin{center}
   \includegraphics[width=0.55\linewidth]{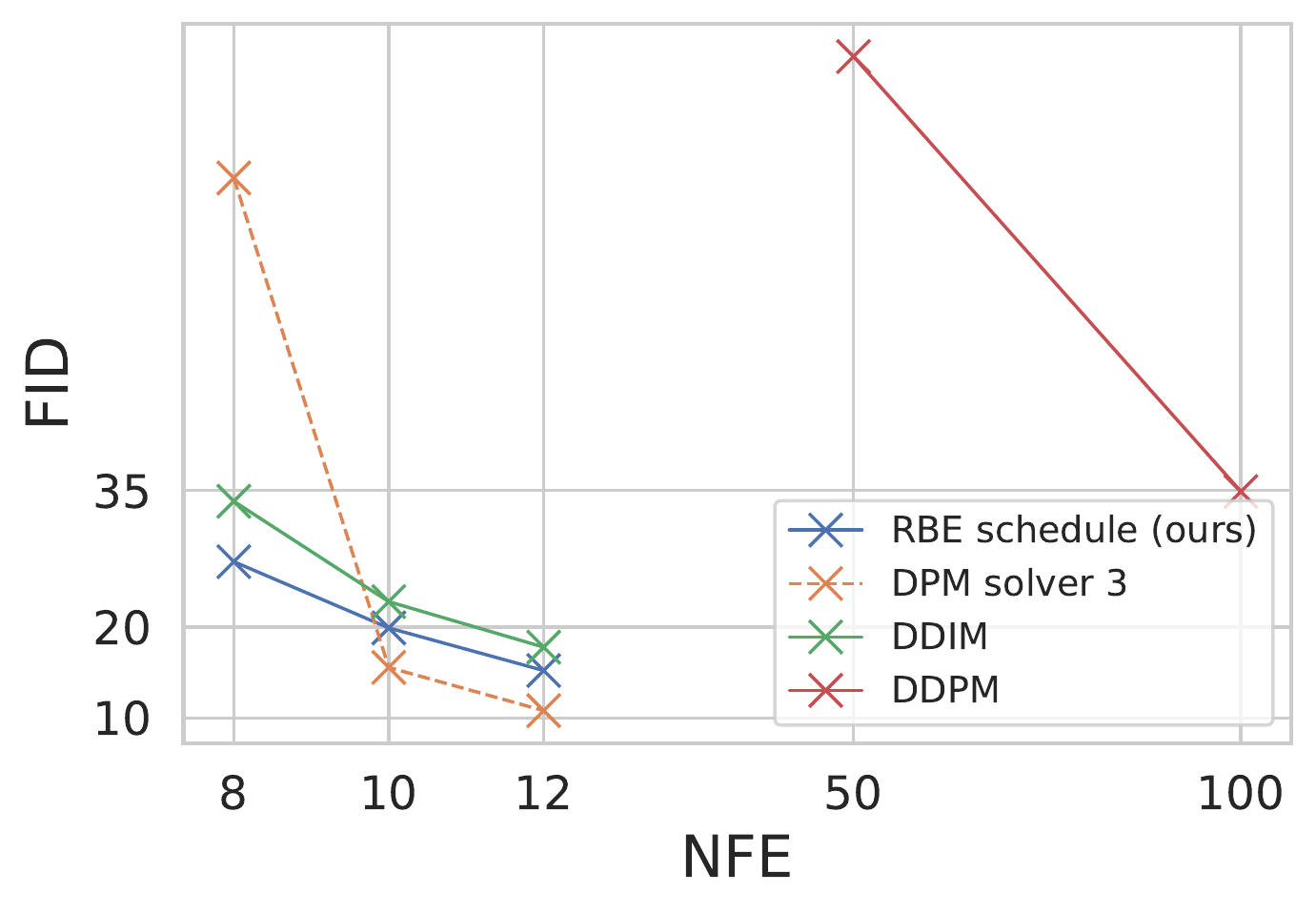}
\end{center}
   \caption{Sample quality measured by FID $\downarrow$ on LSUN bedroom 256×256 with the diffusion model~\cite{nichol2021improved}, varying the number of function evaluations (NFE): 8, 10, 12, 50, 200, 1000. For DDIM and DDPM, we use uniform re-spacing time steps for all the experiments. Sampling through our RBE schedule achieves FID 17.9 in 10 function evaluations and at least 20$\times$ speedup compared with DDPM. }
\label{fig: lsun}
\end{figure}

~\cite{lu2022dpm} evidence that DPM-solvers present the superior efficiency in the few-step regime around 10 to 15 NFE, where other samples such as GGDM~\cite{watson2021learning}, Analytic-DDIM~\cite{bao2022analytic}, exponential integrator~\cite{zhang2022fast} and Itô-Taylor~\cite{tachibana2021taylor} have rather large discretization errors. However, our experiments show that Dpm-solvers~\cite{lu2022dpm} fail dramatically in "fewer" step regimes such as 8 NFE. Meanwhile, sampling through our DEBE and RBE schedule shows the stability and superiority in a broader range of NFEs. As shown in Fig~\ref{fig: img64},~\ref{fig: img128} and~\ref{fig: lsun},  on both ImagesNet and LSUN datasets, sampling through DEBE and RBE schedule can obtain satisfactory samples within 8 steps (FID 29.49 on ImageNet 64x64, FID 12.01 on ImageNet 128x128, and FID 27.18 on LSUN bedroom 256 x 256). Sampling with DRBE and RBE schedule outperforms all the previous samplers in the regime of $8$ inference steps, and achieves comparable performance with the previous fastest training-free samplers such as DPM-solvers~\cite{lu2022dpm} in the regime of 12 to 20 inference steps.


\subsection{ RBE schedule lies between linear and cosine noise schedule}
\begin{figure}[t]
\begin{center}
   \includegraphics[width=0.48\linewidth]{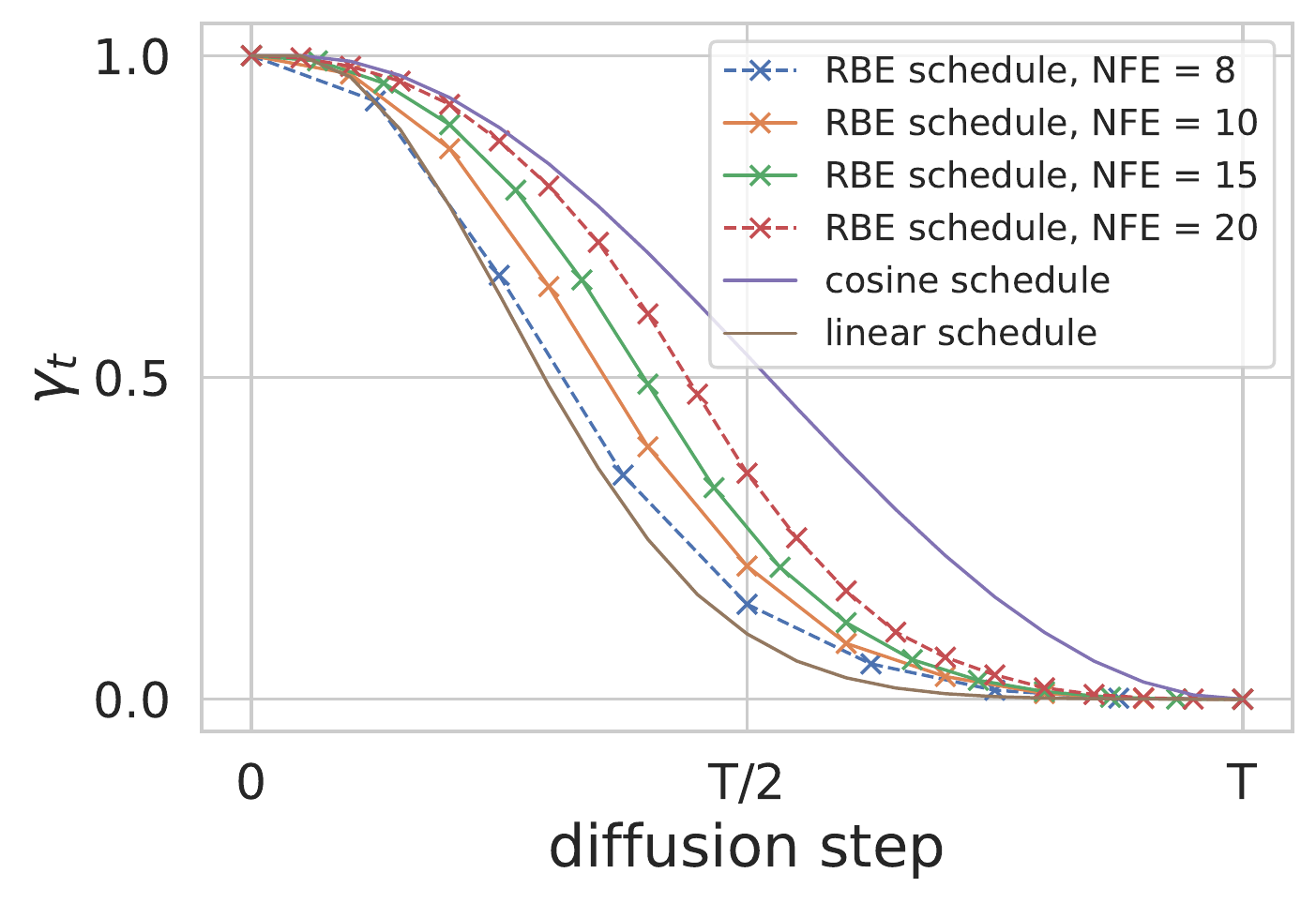}
   \includegraphics[width=0.48\linewidth]{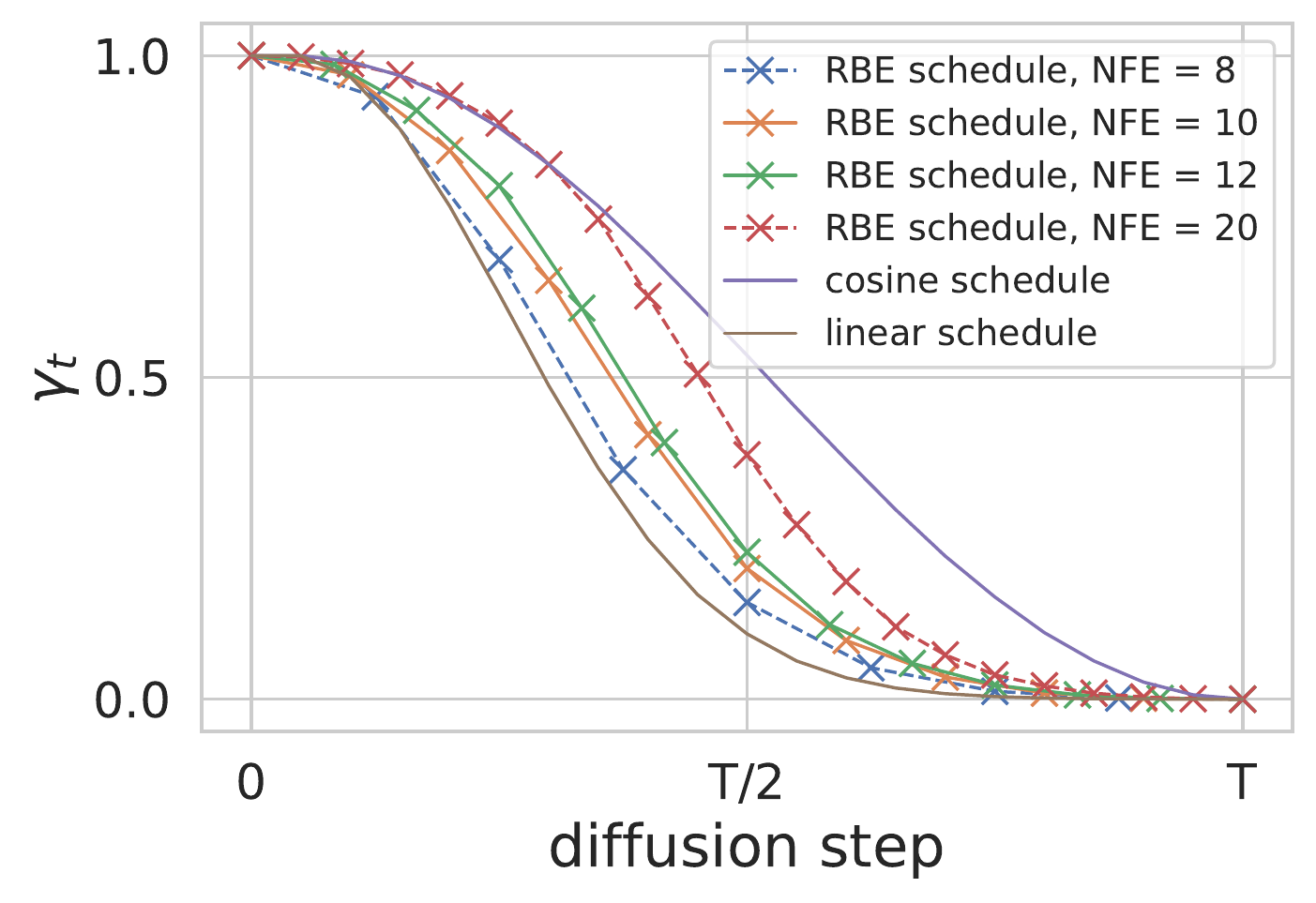}
\end{center}
   \caption{ Comparing our RBE schedule with the linear and cosine schedule. The $x-$axis represents the diffusion steps from $0$ to $T$, while the $y-$axis represents the variance schedule $\g_t$. The left figure plots the RBE schedules of ImageNet 64×64 experiments with the diffusion model~\cite{nichol2021improved}, while the right one presents the RBE schedules of ImageNet 128×128 experiments with the diffusion model~\cite{dhariwal2021diffusion}. RBE schedule is closer to the linear noise schedule in the 8-step regime. Meanwhile, RBE schedule gradually approaches the cosine noise schedule as NFE increases.    }
\label{fig: rbe}
\end{figure}

As derived in section~\ref{sec: RBE}, RBE schedule is learned by restricting the backward error through the sampling process. We have shown that sampling with RBE schedule brings the efficiency of DDPMs samplers to a new level by producing high-quality samples in the “few-step sampling” regime. In this section, we empirically study the behavior of the RBE schedule. In particular, we visualize the RBE schedule in Fig.~\ref{fig: rbe} on both ImageNet 64×64 and ImageNet 128x128 experiments introduced in the previous section. We found that the RBE schedule always lies between the popular choices of noise schedules, such as linear and cosine schedules. RBE schedule is close to the linear noise schedule in the 8 inference steps regime. On the other hand, RBE schedule gradually approaches the cosine noise schedule as NFE increases. This observation motivates us to future research on the design of the optimal noise schedule.

\section{Related work}
\heading{Backward Error Analysis} Backward error analysis is a method for the numerical analysis of ordinary differential equation (ODE) solvers~\cite{hairer2003geometric}. It is a powerful tool for studying the long-time behavior of numerical ODE solvers. In the deep learning community, ~\cite{smith2021origin, barrett2020implicit} analyzed the influence of finite learning rates on the iterates of gradient descent (GD) and shed light on explaining why stochastic gradient descent finds suitable solutions.

\heading{DDPMs samplers} Several works aim at speeding up the inference process for DDPMs. One category of methods includes knowledge distillation~\cite{salimans2022progressive, luhman2021knowledge}, learning sampling trajectory, and searching for better noise schedule~\cite{san2021noise, lam2021bilateral, nichol2021improved, watson2021learning}. Such methods require a training stage before the sampling stage. Hence their applicability and effectivity might be limited when adapting their methods to different models, datasets, and the number of sampling steps. The second category of methods modifies/optimize the forward/backward noising process~\cite{song2020score} in a training-free manner, including dynamic programming~\cite{watson2021learningdp} and analytical generation process~\cite{bao2022analytic}. An important and effective instance is the Denoising Diffusion Implicit Model (DDIM)~\cite{song2020denoising}, a non-Markovian noising process. Another category of training-free methods modifies the numerical solver for stochastic differential equations (SDEs) or ordinary differential equations (ODEs) associated with the DDPMs~\cite{jolicoeur2021gotta, tachibana2021taylor}. In particular, ~\cite{song2020score} solves a marginal equivalent ODE known as the Probability Flow (PF) using a black box ODE solver. ~\cite{liu2022pseudo, lu2022dpm, zhang2022fast} compute DDIM with high order methods and achieve further acceleration. The deterministic DDIM can also be viewed as a time discretization of the PF as it matches the latter in the continuous limit~\cite{song2020denoising}.

.

{\small
\bibliographystyle{ieee}
\bibliography{egbib}
}

\end{document}